\documentclass[journal]{IEEEtran}
\usepackage[utf8]{inputenc} 

\usepackage{algorithm}
\usepackage{algorithmic}
\usepackage{amsmath}
\usepackage{amssymb}
\usepackage{array}
\usepackage{bm}
\usepackage{booktabs}
\usepackage{cases}
\usepackage[pdftex]{graphicx}
\usepackage{indentfirst}
\usepackage{mathrsfs}
\usepackage{multirow}
\usepackage{url}
\usepackage{xcolor}
\usepackage{makecell}
\usepackage{pifont}
\usepackage{color}  
\usepackage{subfigure}
\usepackage{underscore}
\usepackage{soul}
\usepackage{multicol}
\usepackage{multirow}
\usepackage{utfsym}
\usepackage{tablefootnote}
\usepackage[colorlinks,
            linkcolor=blue,
            anchorcolor=black,
            citecolor=green]{hyperref}

\begin{document}

\title{Revisiting RGBT Tracking Benchmarks from the Perspective of Modality Validity: A New Benchmark, Problem, and Method}
\author{Zhangyong~Tang,
        Tianyang~Xu,
        Zhenhua~Feng,
        Xuefeng~Zhu,
        He~Wang,
        Pengcheng~Shao,
        Chunyang~Cheng,
        Xiao-Jun~Wu$^*$,
        Muhammad Awais,
        Sara Atito,
        and Josef Kittler
\thanks{Z. Tang, T. Xu, X. Zhu, H. Wang, P. Shao and X.-J. Wu (Corresponding Author) are with the School of Artificial Intelligence and Computer Science, Jiangnan University, Wuxi, P.R. China. (e-mail: \{zhangyong\_tang\_jnu; tianyang\_xu; xueffeng\_zhu95; 19952730321; pengcheng\_shao; xiaojun\_wu\_jnu\}@163.com)}

\thanks{Z. Feng, M. Awais, S. Atito and J. Kittler are with the Centre for Vision, Speech and Signal Processing, University of Surrey, Guildford, GU2 7XH, UK. (e-mail: \{z.feng; muhammad.awais; sara.atito; j.kittler@surrey.ac.uk)}
}
\maketitle

\begin{abstract}
    RGBT tracking draws increasing attention due to its robustness in multi-modality warranting (MMW) scenarios, such as nighttime and bad weather, where relying on a single sensing modality fails to ensure stable tracking results.
However, the existing benchmarks predominantly consist of videos collected in common scenarios where both RGB and thermal infrared (TIR) information are of sufficient quality. 
This makes the data unrepresentative of severe imaging conditions, leading to tracking failures in MMW scenarios.
To bridge this gap, we present a new benchmark, MV-RGBT, captured specifically in MMW scenarios.
In contrast with the existing datasets, MV-RGBT comprises more object categories and scenes, providing a diverse and challenging benchmark.
Furthermore, for severe imaging conditions of MMW scenarios, a new problem is posed, namely \textit{when to fuse}, to stimulate the development of fusion strategies for such data. 
We propose a new method based on a mixture of experts, namely MoETrack, as a baseline fusion strategy. 
In MoETrack,
each expert generates independent tracking results along with the corresponding confidence score, which is used to control the fusion process.
Extensive experimental results demonstrate the significant potential of MV-RGBT in advancing RGBT tracking and elicit the conclusion that fusion is not always beneficial, especially in MMW scenarios. 
Significantly, the proposed MoETrack method achieves new state-of-the-art results not only on MV-RGBT, but also on standard benchmarks, such as RGBT234, LasHeR, and the short-term split of VTUAV (VTUAV-ST). 
More information of MV-RGBT and the source code of MoETrack will be released at \url{https://github.com/Zhangyong-Tang/MoETrack}.
\end{abstract}


\section{Introduction}
\begin{figure*}[t]
  \centering
  \includegraphics[width=1.0\textwidth]{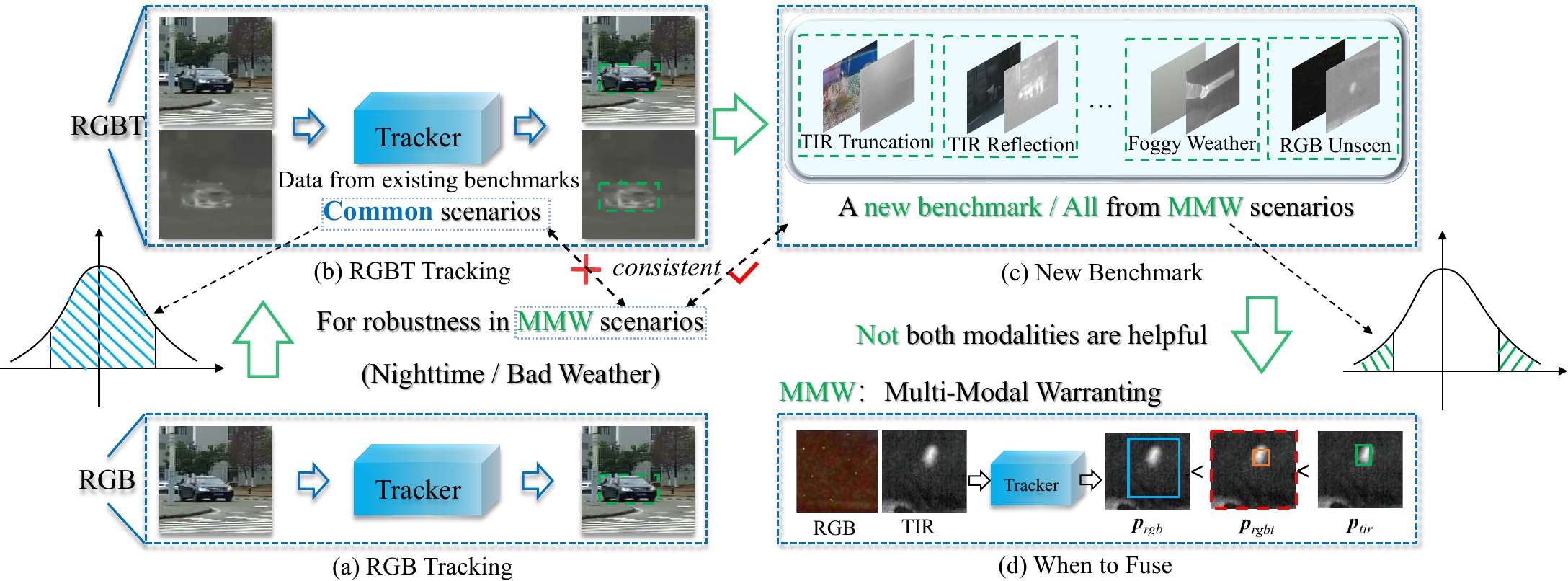}
  \caption{(a) RGB tracking. (b) RGBT tracking. (c) The proposed benchmark inspired by the observed inconsistency between the data in existing benchmarks and the imaging conditions motivating RGBT tracking. (d) Taking into account the modality validity, a new problem \textit{when to fuse} is posed in this work, which discusses strategies for multi-modality information fusion. `MMW' is the abbreviation of `multi-modality warranting'. $\mathbf{\textit{p}}_{rgb}$, $\mathbf{\textit{p}}_{tir}$, and $\mathbf{\textit{p}}_{rgbt}$ represent the predictions produced by the RGB, TIR, and the fused (RGBT) experts, respectively.}
  \label{fig:difference}
\end{figure*}

Visual object tracking is a hot topic in computer vision, aiming to predict the location and size of an object throughout a video sequence, starting from its initial state specified in the first frame \cite{vot2023, xu2020accelerated, xu2023learning}.
Recent studies have identified the limitations of using only visible sensors, leading to a growing interest in integrating auxiliary modalities such as thermal infrared (TIR) \cite{lasher}, depth \cite{rgbd1k}, event \cite{coesot}, and language \cite{unimod1k}. 
This trend has propelled multi-modality tracking into the spotlight.
RGBT tracking, in particular, has emerged as a popular topic due to the complementary characteristics of RGB and TIR modalities.
For instance, RGB data is sensitive to changing illumination conditions, whereas TIR data is not \cite{mufusion}. 
Conversely, TIR data lacks colour information which is typically contained in RGB data \cite{textfusion}. 
In other words, compared to the reliance on a single modality, RGBT tracking offers distinct advantages, helping to stabilise the tracking, especially when one modality encounters significant challenges, such as thermal crossover and over-exposure. These severe imaging conditions are referred as multi-modality warranting (MMW) scenarios in this work.

Thanks to the rapid development of RGB and TIR sensors, various RGBT tracking benchmarks have been proposed, such as GTOT \cite{gtot}, RGBT210 \cite{RGBT210}, RGBT234 \cite{rgbt234}, LasHeR \cite{lasher}, and VTUAV \cite{hmft}, accelerating the research in the domain.
However, a statistical analysis of these benchmarks,  which involves sampling 20\% of the videos at random to determine whether they are captured under MMW scenarios or not, indicates that \textit{almost all the videos are collected under common scenarios, presenting no critical imaging condition challenges}.
This observation indicates that these benchmarks are unrepresentative of MMW scenarios and by implication, the advantages of combining RGB and TIR modalities have not been fully investigated.
Furthermore, the robustness of existing methods in MMW scenarios remains unexplored, leading to unreliable recommendations when deploying RGBT trackers in practical applications.

To address the above issues and rectify the deficiencies of the current benchmarks, we propose a new benchmark for RGBT tracking, which solely contains data collected in MMW scenarios.
Since one modality is usually non-informative in MMW scenarios, as exemplified in Figure \ref{fig:difference}(c), the proposed benchmark, MV-RGBT, aims to draw more attention to modality validity.
Essentially, MV-RGBT can be futher divided into two subsets: MV-RGBT-RGB and MV-RGBT-TIR.
For example, the RGB modality is unseen in the nighttime, and such videos belong to MV-RGBT-TIR since the TIR modality provides unaffected perception of target, and vice versa.
This categorisation allows us to consider the effectiveness of the solutions in a compositional manner, enabling an in-depth analysis and providing insights for future developments.
More discussions are provided in Sec. \ref{sec:modalitybias}.

Furthermore, the frequent presence of non-informative data in MMW scenarios prompts us to delve into the necessity of multi-modality information fusion, posing the problem of \textit{when to fuse}, as aggregating irrelevant data may be unhelpful or even harmful.
While designing a classifier to gauge data validity at the image level might be the most straightforward solution, 
the scarcity of data for training such classifiers precludes this option.
Consequently, our approach focuses on the decision level. We note that non-informative data tends to produce coarse bounding box predictions under the widely-used tracking-by-detection paradigm, as illustrated in Figure. \ref{fig:difference}(d). At the same time, one of the imaging modalities may be competent in tracking the target on its own. 
Thus, our approach generates in parallel separated RGB, TIR, and fused modality predictions.
Methodologically, the proposed approach  deploys a \textbf{M}ixture \textbf{o}f \textbf{E}xperts, including the RGB, TIR, and RGBT experts, dubbed as MoETrack.
During inference, each expert provides a bounding box prediction along with the corresponding confidence score.
The final prediction is controlled by the confidence score, which determines \textit{when to fuse}.
Specifically, if the RGBT expert produces the highest score, the corresponding bounding box results will be selected, indicating that fusion is considered beneficial and vice versa.

In summary, the main contributions of this work include:

$\bullet$ A new benchmark dataset, MV-RGBT, is collected to make it representative of MMW scenarios.
Furthermore, videos in the dataset can be categorised according to the modality informativeness into two subsets, MV-RGBT-RGB and MV-RGBT-TIR,  facilitating an in-depth analysis of the tracking methods in a compositional way.

$\bullet$ A new problem, \textit{when to fuse}, is introduced to develop a reliable fusion strategy, as in MMW scenarios multi-modality information fusion may be counterproductive. 

$\bullet$ A new fusion method, MoETrack, is proposed,  involving three tracking heads (experts). It offers a more flexible way to deal with both the RGB- and TIR-specific challenges by adaptively switching the tracking prediction to the one from the most reliable expert.

$\bullet$ Extensive experiments demonstrate that MoETrack defines new state-of-the-art results on several benchmarks, including MV-RGBT, RGBT234, LasHeR, and VTUAV-ST.

\begin{table*}[tb]
  \caption{A comparison between existing RGBT tracking benchmarks and the proposed MV-RGBT benchmark.
  }
  \label{tab:benchmarks}
  \centering
  \small
  \begin{tabular}{ccccccccc}
    \toprule
    \multirow{2}{*}{Benchmarks} & Num. & Avg. & Max. & Total & Modality & \multirow{2}{*}{Resolution} & \multirow{2}{*}{Category} & \multirow{2}{*}{Scene}  \\
    & Seq. & Frame & Frame & Frame &  Validity & & & \\
    \midrule
    GTOT \cite{gtot} & 50 & 157 & 376 & 7.8k & \usym{2717} & 384x288 & 9 & 6\\
    RGBT210 \cite{RGBT210} & 210 & 498 & 4140 & 104.7k & \usym{2717} & 630x460 & 22 & 8 \\
    RGBT234 \cite{rgbt234} & 234 & 498 & 4140 & 116.7k & \usym{2717} & 630x460 & 22 & 8 \\
    VOT-RGBT2019 \cite{VOT2019} & 60 & 334 & 1335 & 40.2k & \usym{2717} & 630x460 & 13 & 5 \\
    VOT-RGBT2020 \cite{VOT2020} & 60 & 334 & 1335 & 40.2k & \usym{2717} & 630x460 & 13 & 5 \\
    LasHeR-test \cite{lasher} & 234 & 900 & 12862 & 22.0k & \usym{2717} & 630x460 & 19 & 15 \\
    VTUAV-test-st \cite{hmft} & 176 & 3588 & 25295 & 63.1k & \usym{2717} & 1920x1080\footnotemark[1] & 13 & 10 \\
    \midrule
    Ours(MV-RGBT) & 122 & 737 & 2113 & 89.9k & \usym{1F5F8} & 640x480 & 36 & 19 \\
  \bottomrule
  \end{tabular}
\end{table*}

\begin{figure*}[tb]
  \centering
  \includegraphics[width=0.9\textwidth]{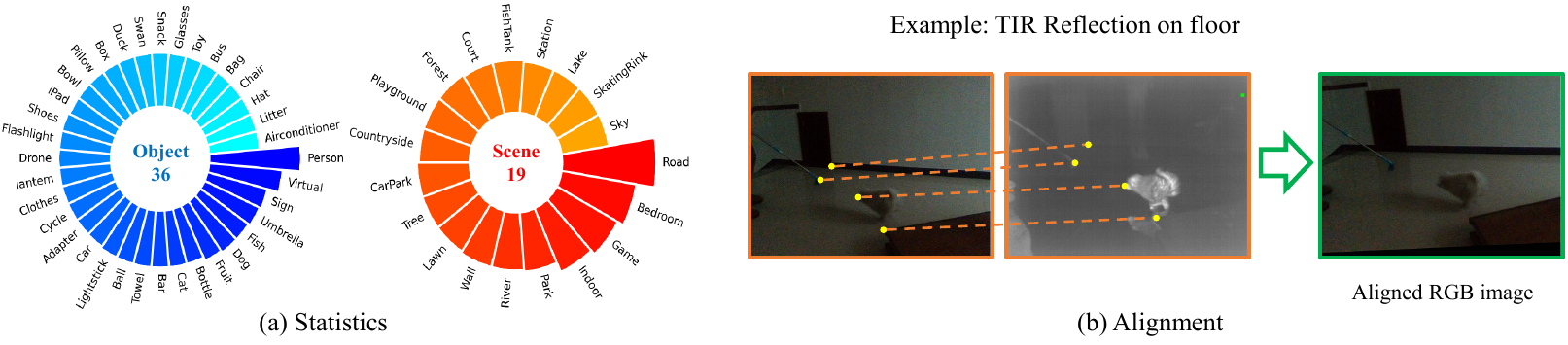}
  \caption{(a) The statistics of the MV-RGBT benchmark and (b) a brief introduction of the key-point-based alignment.}
  \label{fig:example}
\end{figure*}

\section{Related Work}
\subsection{RGBT Tracking Benchmarks}
With the popularity of RGB and TIR sensors, several RGBT tracking benchmarks have been proposed.
As shown in Table \ref{tab:benchmarks}, there are 7 popular RGBT tracking benchmarks, including GTOT \cite{gtot}, RGBT210 \cite{RGBT210}, RGBT234 \cite{rgbt234}, LasHeR \cite{lasher},  VTUAV \cite{hmft}, VOT-RGBT2019 \cite{VOT2019}, and VOT-RGBT2020 \cite{VOT2020}.
Among them, GTOT \cite{gtot} stands out as the pioneering benchmark, comprising 50 video pairs.
However, in the deep learning era, this is of a very limited size for comprehensive evaluation.
To address this, RGBT210 \cite{RGBT210} is proposed with 210 video pairs.
Furthermore, it is extended by including videos in more scenes, such as hot days, forming a new benchmark named RGBT234 \cite{rgbt234}.
VOT-RGBT2019 \cite{VOT2019} and VOT-RGBT2020 \cite{VOT2020} are subsets of RGBT234, but with different benchmarking strategies.
VOT-RGBT2019 employs a re-start strategy when a tracking failure is detected.
However, during evaluation, the location, where the tracker fails, affects the final performance, which is considered unreasonable.
Hence, the re-start strategy is replaced in VOT-RGBT2020 with a multi-start strategy, executing the tracker from multiple fixed anchors.
LasHeR \cite{lasher} is another benchmark proposed together with a large-scale training set and contains 245 video pairs captured in 15 scenes.
Unlike the above benchmarks collected from a human or surveillance perspective, VTUAV \cite{hmft} collects data from an unmanned aerial vehicle (UAV).
Its short-term split consists of 176 video pairs, with an average length of 3500 frames.

However, as shown in Table \ref{tab:benchmarks}, it is evident that the data from the aforementioned benchmarks are predominantly collected in common scenarios, which markedly differ from the multi-modality warranting (MMW) scenarios discussed when highlighting the advantages of RGBT tracking.
Our MV-RGBT benchmark bridges this gap by ensuring that all the videos are collected in MMW scenarios.
Additionally, based on the specific challenges unique to each modality, MV-RGBT is divided into RGB and TIR components. 
This division allows for a detailed analysis in a compositional manner, facilitating a more comprehensive assessment of the contribution of each modality and their fusion for more nuanced deployment of RGBT trackers.

\subsection{Multi-Modality Information Fusion}
As a key element in RGBT tracking, the fusion of the multi-modality information is always crucial for a high-performance tracker.
According to the location where the fusion happens, existing fusion strategies can be divided into pixel- \cite{pixel-fusion2007}, feature- \cite{MANet, mfDimp, FANet, cbpnet, cat++, bat, tbsi, vipt, gmmt}, and decision-level \cite{dfat, taat, hmft} methods.
Pixel-level fusion involves a fusion in the image domain, necessitating perfectly aligned multi-modality data. 
However, this constraint limits the attention of the research community paid to this level of fusion. 
Consequently, feature-level and decision-level fusion methods have become more popular.
At the feature level, information fusion is implemented at a high-dimensional semantic space, offering a more comprehensive multi-modality perception at the expense of increased computational resources \cite{apfnet, MANet, cbpnet, FANet, gmmt, bat}.
On the other hand, decision-level fusion strategies typically utilise the intermediate tracking results produced by each modality, embedded in a low-dimensional task-related space \cite{dfat, hmft}. 
Compared to the trackers relying on feature-level fusion, decision-level fusion methods often exhibit higher efficiency, while maintaining comparable performance.

However, regardless of the level at which the fusion block is placed, existing methods integrate multi-modality information at every frame.
Despite their promising performance, there has been a notable lack of discussion of this strategy. 
We argue that this strategy is sub-optimal in Sec. \ref{sec:experiment}.
For example, qualitatively, in MMW scenarios, one of the modalities often encounters severe challenges, making it non-informative, and potentially even causing the injection of harmful information.
In such situations, the adoption of standard fusion strategy warrants further assessment. 
Therefore, a new problem \textit{when to fuse} is addressed to enhance the effectiveness of multi-modality information fusion.

In the light of this, we develop a new RGBT tracking method with multiple tracking heads.
Each of them provides a distinct prediction and therefore they act as different experts, referred to as RGB, TIR, and RGBT experts, based on their inputs.
Subsequently, an expert selection strategy forms our decision-level fusion block, with the final choice opting for fusion-based tracking only if fusion is deemed crucial.

\begin{figure*}[t]
  \centering
  \includegraphics[width=1.0\textwidth]{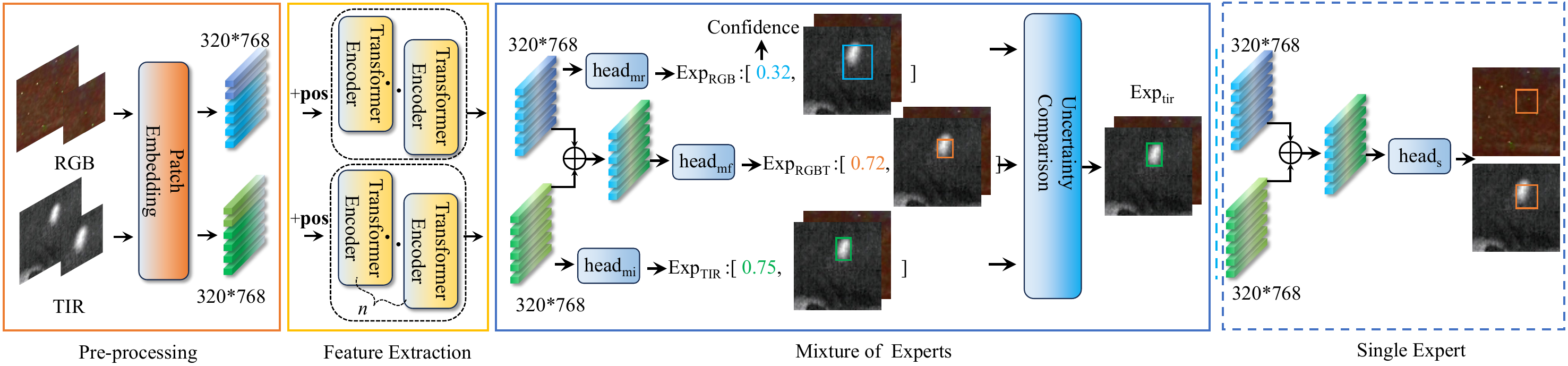}
          \caption{Pipeline of the proposed MoETrack. Considering three experts, RGB, TIR, and RGBT, a comparison of confidence scores is conducted to maintain the final prediction.} 
  \label{fig:pipeline}
\end{figure*}

\footnotetext[1]{It is obtained by super-resolution.}

\section{New Benchmark: MV-RGBT}
\subsection{Data Collection}\label{sec:collection}
Since the objective is to address the inconsistency between the data in current benchmarks and MMW scenarios, where the utilisation of multi-modal data is critical for a stable tracking system, MV-RGBT is totally captured in MMW scenarios.
As depicted in Figure \ref{fig:difference}(c), our fundamental idea is to identify MMW scenarios.
In MMW scenarios, one modality typically suffers significant challenges specific to its physical properties \cite{lasher}, while the other remains relatively unaffected.
Therefore, in MV-RGBT, the challenges are categorised as RGB-specific and TIR-specific:
(1) Bad weather: This refers to conditions where the visibility of RGB channels is severely impacted, such as heavy foggy days.
(2) Extreme illumination: This occurs when objects are not visible during nighttime or overexposure. 
(3) TIR truncation: TIR radiation is unable to penetrate transparent objects, such as water surfaces or glass.
(4) TIR reflection: Coexistence of different TIR radiations for the same objects, especially when objects are near reflective surfaces like mirrors. 
(5) TIR background clutter: Inanimate objects that remain in the same space for an extended period tend to blend with the environment, such as the umbrellas used outdoors on rainy days.  

Following the aforementioned principles, a platform equipped with a TIR sensor (FLIR BOSON PlUS 640 \footnotemark[2]) and an RGB sensor (Intel RealSence Depth camera D456 \footnotemark[3]) is assembled for data collection.
As shown in Table \ref{tab:benchmarks}, MV-RGBT comprises 122 multi-modal video pairs.
The average and maximum video lengths are 737 and 2113 frames, respectively.
In total, MV-RGBT consists of 89.9k frame pairs with a resolution of 640x480.
Furthermore, the objects in MV-RGBT come from 36 different classes, and the videos are captured in 19 distinct scenes, rendering them more diverse than the other publicly available benchmarks. 
Specific details are shown in Figure \ref{fig:example}(a).
\footnotetext[2]{https://www.flir.co.uk/products/boson-plus/}
\footnotetext[3]{https://www.intelrealsense.com/}

\subsection{Data Annotation and Alignment}
MV-RGBT benefits from meticulous annotation efforts by several researchers in the field of visual object tracking.
Notably, the provided rectangle-formatted annotations strictly enclose only the visible parts of objects.
In cases where objects are completely unseen or occluded, all values of rectangle are set to 0.
For the alignment of different modalities, the widely recognised key-point-based algorithm, LoFTR \cite{loftr}, is employed. 
However, when LoFTR fails to provide satisfactory results, manually annotated key points are utilised, as depicted in Figure \ref{fig:example}(b), ensuring accurate alignments between different modalities of each frame. 
Ultimately, the entire MV-RGBT benchmark undergoes strict quality checks to ensure high-quality annotations throughout.

\subsection{Evaluation Metrics}
\label{sec:metrics}

\begin{equation}
    \centering \label{eq:metrics}
    \begin{aligned}
\rm sr & = \frac{1}{m} \sum_{j=1}^{m}\left(\frac{1}{t} \sum_{i=1}^{t} \rm{IoU}(\mathit{\textbf{\textit{g}}_{j, i}, \textbf{\textit{p}}_{j, i}}) > th_s\right) \\
\rm pr & = \frac{1}{m} \sum_{j=1}^{m}\left(\frac{1}{t} \sum_{i=1}^{t} \rm{Dis} (\mathit{\textbf{\textit{g}}_{j, i, c} , \textbf{\textit{p}}_{j, i, c}}) > th_p\right)
    \end{aligned}
\end{equation}

Similar to other tracking benchmarks, as presented in Eq. \eqref{eq:metrics}, the intersection over union (IoU) between the ground truth bounding box $\textbf{\textit{g}}_{j, i}$ and predicted bounding box $\textbf{\textit{p}}_{j, i}$ is calculated for evaluation, as well as the $\ell$2 distance (Dis) between the centre of these bounding boxes, $\textbf{\textit{g}}_{j, i, c}$ and $\textbf{\textit{p}}_{j, i, c}$.
The subscript $j$ denotes the $j$-th video and $i$ means the index of the frame.
$c$ signifies `centre'.
$t$ and $m$ denote the frames contained in the sequence and videos contained in the entire benchmark, respectively.
$\rm th_s$ and $\rm th_p$ represent the thresholds for calculating the success rate $\rm sr$ and precision rate $\rm pr$.
That means IoU and the centre distance are first averaged across all frames within each sequence, and then across all sequences. 
Later, in order to provide a comprehensive evaluation, multiple thresholds are employed and the results under each threshold are recorded.
Consequently, the area under curve (AUC) is reported as the final score, which is displayed in Figure \ref{fig:results}.
Notably, for absent frames, if there exists a prediction, the outputs of the IoU($\cdot$) and Dis($\cdot$) functions are both 0. However, if the absence is predicted, these outputs are set to 1.

\section{New Solution: MoETrack}
\subsection{RGBT Tracking}
Given the $i$-th multi-modality frame pair $\textbf{\textit{X}}_{i, rgb}$ and $\textbf{\textit{X}}_{i, tir}$, the goal of an RGBT tracker is to obtain the bounding box prediction of the current frame:
\begin{equation}
    \centering \label{eq:mmtrack}
    \begin{aligned}
\textbf{\textit{p}}_i= f(\textbf{\textit{X}}_{i, rgb}; \textbf{\textit{X}}_{i, tir}; \boldsymbol{\theta}; \boldsymbol{\phi}), \\
    \end{aligned}
\end{equation}
where $f(\cdot)$ denotes the tracker with offline-learned parameters $\boldsymbol{\theta}$.
Notably, $\boldsymbol{\phi}$ represents the weights used for multi-modality information fusion, which is typically employed in every frame.

\subsection{MoETrack}
After collecting the data in MMW scenarios, as depicted in Figure \ref{fig:difference}(c), the loss of information in one modality inspires us to reconsider the necessity of fusion, leading to our discussion on \textit{when to fuse}. 
In response, MoETrack is developed with multiple tracking heads with each of them acting as an expert.
The appropriate selection of one of  these experts generates the best prediction of the tracked object.

\textbf{Network Overview:}
As illustrated in Figure \ref{fig:pipeline}, the multi-modality frame pair $\textbf{\textit{X}}_{i, rgb}$ and $\textbf{\textit{X}}_{i, tir}$ are firstly divided into patches and then transferred into tokens.
Since the spatial structure is broken during tokenisation, a learnable positional embedding is further introduced, whose outputs $\textbf{\textit{X}}^{pe}_{i, rgb}$ and $\textbf{\textit{X}}^{pe}_{i, tir}$ $\in \mathbb R^{k\times d}$ form the inputs of the transformer-based backbone, where $k$ is the number of tokens and $d$ denotes the length of each token.
As to the backbone, the ViT-base provided by \cite{ostrack} is employed, containing $n=12$ standard transformer encoders.
Additionally, for both RGB and TIR branches, the backbone is shared and their corresponding outputs $\textbf{\textit{X}}^b_{i, rgb}$ and $\textbf{\textit{X}}^b_{i, tir}$ $\in \mathbb R^{k\times d}$ also share the same dimensions.
After that, they are combined as the fused feature $\textbf{\textit{X}}^b_{i, rgbt}$ $\in \mathbb R^{k\times d}$, which is then transferred into the task-related space through a tracking head. 
However, it merely acts like the RGBT expert in our design of mixture of experts and the fusion is simply defined by an element-wise addition.
Hence, other two tracking heads are adopted for $\textbf{\textit{X}}^b_{i, rgb}$ and $\textbf{\textit{X}}^b_{i, tir}$, serving as the RGB and TIR experts, respectively.
Later, the final prediction $\textbf{\textit{p}}_i$ is provided by the expert with the highest confidence score.
Notably, the maximum score of the classification map serves as a confidence measure in our work, as it is widely used as a reliability measurement in the tracking field \cite{dfat, hmft}.

In this manner, with adaptive selection implemented, the RGBT tracking paradigm introduced in Eq. \eqref{eq:mmtrack} evolves into a new one:
\begin{align} \label{eq:moetrack}
    \textbf{\textit{p}}_i=\left\{
    \begin{aligned}
    & f(\textbf{\textit{X}}_{i, rgb}; \boldsymbol{\theta}), \text{if} \; \rm{mc} = \mathit{cs_{rgb}}; \\
    & f(\textbf{\textit{X}}_{i, rgb}; \textbf{\textit{X}}_{i, tir}; \boldsymbol{\theta}; \boldsymbol{\phi}), \text{if} \; \rm{mc} = \mathit{cs_{rgbt}}; \\
    & f(\textbf{\textit{X}}_{i, tir}; \boldsymbol{\theta}), \text{if} \; \rm{mc} = \mathit{cs_{tir}}; \\
\end{aligned}
\right.
\end{align}
where $cs_{rgb}$, $cs_{tir}$, and $cs_{rgbt}$ denote the confidence scores of RGB, TIR, and RGBT experts, respectively.
$\rm mc = \max({\textit{cs}_{\textit{rgb}}, \textit{cs}_{\textit{tir}}, \textit{cs}_{\textit{rgbt}}})$ is obtained as the maximum score.
Therefore, the selection should intuitively reflect whether fusion is necessary or not, which naturally supports our further discussion on \textit{when to fuse}.

\textbf{Network Training:} In our design, the backbone is finetuned and the other parameters are trained from scratch according to the gradients from multiple experts.
Basically, each expert is assigned a tracking loss $l$ to ensure specialisation and $l$ is calculated by following \cite{vipt}.
The final loss is computed by averaging all the expert losses:
\begin{align} \label{eq:loss}
    \begin{aligned}
    loss = (l_{rgb} + l_{tir} + l_{rgbt})/3, \\
\end{aligned}
\end{align}
where $l_{rgb}$, $l_{tir}$, and $l_{rgbt}$ represent the loss for RGB, TIR, and RGBT experts, respectively.

\begin{table*}[t]
  \caption{Quantitative results on GTOT, RGBT234, LasHeR, and VTUAV-ST.
  }
  \centering \label{tab:results}
  \resizebox{1\linewidth}{!}{
\begin{tabular}{cccccccccccccccccc}
\toprule
\multicolumn{1}{c}{\multirow{2}{*}{Method}} & \multirow{2}{*}{Venue} & \multicolumn{3}{c}{GTOT}  & & \multicolumn{3}{c}{RGBT234} & & \multicolumn{3}{c}{LasHeR} & &  \multicolumn{3}{c}{VTUAV-ST}  &  FPS\\ 
\multicolumn{1}{c}{} & & PR/\% $\uparrow$ & & SR/\% $\uparrow$& & PR/\% $\uparrow$ & & SR/\% $\uparrow$& & PR/\% $\uparrow$ & & SR/\% $\uparrow$& & PR/\% $\uparrow$& & SR/\% $\uparrow$ & $\uparrow$\\
\midrule
FSRPN \cite{VOT2019}  & ICCVW2019 & 89.0& &69.5 & & 71.9 & & 52.5 & & - & & - & &  65.3& &54.4 & -\\
mfDiMP \cite{mfDimp}  & ICCVW2019 & 83.6& &69.7 & & 84.6 & & 59.1 & & 44.7 & & 34.3 & &  67.3& &55.4 & 10.0\\
DAFNet \cite{DAFNet}  & ICCVW2019 & 89.1& &71.6 & & 79.6 & & 54.4 & & 44.8 & & 31.1 & &  62.0& &45.8 & 20.0\\
CAT \cite{cat} & ECCV2020 & 88.9& &71.7 & & 80.4 & & 56.1 & & 45.0 & & 31.4 & & - & & -& 20.0\\
CMPP \cite{cmpp} & CVPR2020 & 92.6& &73.8 & & 82.3 & & 57.5 & & - & & - & & - & & -& 1.3\\
MANet++ \cite{manet++} & TIP2021& 88.2& &70.7 & & 80.0 & & 55.4 & & 46.7 & & 31.4 & & - & &- & 25.0\\
JMMAC \cite{JMMAC} & TIP2021& 90.2& &73.2 & & 79.0 & & 57.3 & & 46.7 & & 31.4 & & - & &- & 4.0\\
ADRNet \cite{adrnet} & IJCV2021& 90.4& &73.9 & & 80.7 & & 57.0 & & - & & - & & 62.2 & &46.6 & 25.0\\
MFGNet \cite{mfgnet} & TMM2022& 88.9& &70.7 & & 78.3 & & 53.5 & & - &  & - & & - & & -& -\\
DMCNet \cite{DMCNet} & TNNLS2022& 90.9& &73.3 & & 83.9 & & 59.3 & & 49.0 & & 35.5 & & - & &- & 2.3\\
APFNet \cite{apfnet} & AAAI2022& 90.5& &73.7 & & 82.7 & & 57.9 & & 50.0 &  & 36.2 & & - & & -& 1.3\\
ProTrack \cite{protrack} & ACMMM2022& -& &- & & 78.6 & & 58.7 & & 50.9 &  & 42.1 & & - & &- & 30.0\\
MIRNet \cite{mirnet} & ICME2022& 90.9& &74.4 & & 81.6 & & 58.9 & & - &  & - & & - & & -& 30.0\\
HMFT \cite{hmft}  & CVPR2022 & 91.2& &74.9 & & 78.8 & & 56.8 & & - & & - & & 75.8 & &62.7 & 30.2\\
LANet \cite{lanet} & TMM2023& 91.3& &75.1 & & 79.5 & & 58.4 & & 53.8 & & 43.1 & & - & &- & 21.7\\
ECMD \cite{ecmd}  & CVPR2023 & 90.7& &73.5 & & 84.4 & & 60.1 & & 59.7 & & 46.7 & & - & & -& 30.0\\
QAT \cite{qat} & ACMMM2023 & 91.5& &75.5 & & \textcolor{red}{\textbf{88.4}} & & 64.3 & & 64.2 & & 50.1 & & 80.1 & &66.7 & 22.0\\
ViPT \cite{vipt}  & CVPR2023 & 91.4& &76.3 & & 83.5 & & 61.7 & & 65.1 & & 52.5 & & - & & -& \textcolor{red}{\textbf{39.0}}\\
TBSI \cite{tbsi} & CVPR2023 & 91.5& &75.9 & & 87.1 & & 63.8 & & 69.2 & & 55.6 & & - & & -& 36.0\\
CAT++ \cite{cat++} & TIP2024 & 91.5& &73.3 & & 84.0 & & 59.2 & & 50.9 & & 35.6 & & - & & -& 14.0\\
BAT \cite{bat} & AAAI2024 & 90.9& &76.3 & & 86.8 & & 64.1 & & 70.2 & & 56.3 & & 81.8 & &67.4 & 15.0\\
GMMT \cite{gmmt} & AAAI2024 & \textcolor{red}{\textbf{93.6}}& &\textcolor{red}{\textbf{78.5}} & & 87.9 & & 64.7 & & 70.7 & & 56.6 & & 82.8 & &68.5 & 20.0\\
\midrule
SETrack&  & 91.7& &76.6 & & 87.1 & & 64.4 & & 71.7 & & 57.2 & & 82.7& &68.7 & 25.0\\
MoETrack-TIR& & 64.3 & & 56.3 & & 76.5& &54.0 & & 59.8 & & 47.4 & & 51.7 & &41.2 & 25.0\\
MoETrack-RGB& & 84.9& &68.9 & & 81.6 & & 60.7 & & 62.4 &  & 50.2 & & 76.1 & & 65.7 & 25.0\\
MoETrack-RGBT & & 92.9& &77.7 & & 87.5 & & 64.8 & & 71.7 &  & 57.5 & & 82.9 & &69.1 & 25.0\\
MoETrack & & \textcolor{red}{\textbf{93.6}}& &78.4 & & 88.1 & & \textcolor{red}{\textbf{65.1}} & & \textcolor{red}{\textbf{72.1}} & & \textcolor{red}{\textbf{57.8}} & & \textcolor{red}{\textbf{83.6}} & &\textcolor{red}{\textbf{69.5}} & 23.0\\
\bottomrule
\end{tabular}
  
  }
\end{table*}

\section{Experiments} \label{sec:experiment}
\subsection{Implementation Details}
The implementation of our MoETrack is executed on a platform equipped with an NVIDIA RTX 3090Ti GPU card. 
ViT-256 is employed as the backbone and it is finetuned by AdamW with gradients learned from the training split of LasHeR \cite{lasher}.
The learning rate is initialised at 7.5e-5 and drops to one-tenth of the current value for every 10 epochs.
The batch size is set to 32.

\subsection{Evaluation Data and Metrics}

The effectiveness of the proposed method, MoETrack, is verified on the proposed MV-RGBT benchmark as well as several publicly available benchmarks, including GTOT \cite{gtot}, RGBT234 \cite{rgbt234}, LasHeR \cite{lasher}, and VTUAV \cite{hmft}.
Since the MV-RGBT is thoroughly introduced in the main file, the details of other benchmarks are displayed in the following paragraphs.

GTOT is an early published RGB-T dataset, including 7.8K image pairs.
The evaluation metrics are precision rate (PR) and success rate (SR).
PR measures the percentage of frames with the distance between centres of the predicted and ground truth bounding box below a threshold, 5 in this benchmark.
SR represents the ratio of frames being tracked with the overlap between the predicted and ground truth bounding box above zero.

RGBT234 contains 234 and 116.7K multi-modal video and frame pairs, respectively.
This benchmark employs the same evaluation metrics with GTOT.

LasHeR is a large and widely-used benchmark in the RGB-T field, and its testing split consists of 245 video pairs.
PR, SR and the normalised precision rate (NPR) are used for benchmarking.
NPR \cite{trackingnet} is a modified version of PR since PR can be easily affected the image resolution and the size of the ground truth bounding box.
It should be noted that the threshold of PR in LasHeR is 20.
However, only PR ans SR are involved in this work, as the comparison of PR always produces the same conclusion with that of NPR.

VTUAV is a benchmark collected by drones.
Basically, it has both the long-term and short-term splits.
However, since almost all the competitive methods are short-term trackers, only the comparison on the short-term split (VTUAV-ST) is provided in the main file.
VTUAV-ST contains 176 videos, with an average length of 3500 frames.
On this benchmark, we also use the PR and SR as evalution metrics.

\subsection{Quantitative Analysis}
To provide a comprehensive evaluation of our method, the experiments are carried out on our MV-RGBT and four existing benchmarks, including GTOT \cite{gtot}, RGBT234 \cite{rgbt234}, LasHeR \cite{lasher}, and VTUAV-ST \cite{hmft}.
We compare MoETrack with 22 advanced trackers in Table~\ref{tab:results}. According to the table, MoETrack demonstrates promising performance on these benchmarks and new state-of-the-art records are established on RGBT234, LasHeR, VTUAV-ST, and MV-RGBT.

As illustrated in Table \ref{tab:results}, on GTOT, our method achieves PR and SR results of 93.6\% and 78.4\%, respectively. 
Compared to the best-performing tracker GMMT \cite{gmmt}, our method exhibits the same performance on PR and a slight degradation (0.1\%) on SR.
On RGBT234, our method performs the best on SR (65.1\%) and the second on PR (88.1\%).
Regarding LasHeR, again, our method ranks first on PR and SR, achieving 72.1\% and 57.8\%, respectively.
To further demonstrate the generalisation capability of our method, evaluations on VTUAV-ST are conducted, as its data is collected from a different perspective compared to the training set \cite{hmft}.
On VTUAV-ST, our method performs the best in terms of PR and SR, reaching 83.6\% and 69.5\%, respectively.

Furthermore, according to Table \ref{tab:results}, several advanced trackers are included for benchmarking on our benchmark, MV-RGBT.
The results are reported in Figure \ref{fig:results} (upper figures) and our method shows more recognisable advantages.
Specifically, the results on PR and SR are 51.4\% and 67.6\%, respectively.
Compared to the second-place tracker, GMMT, our method has improvements of 2.3\% and 2.2\% on PR and SR, highlighting the effectiveness of our method.

\begin{figure*}[t]
  \centering
  \includegraphics[width=1.0\textwidth]{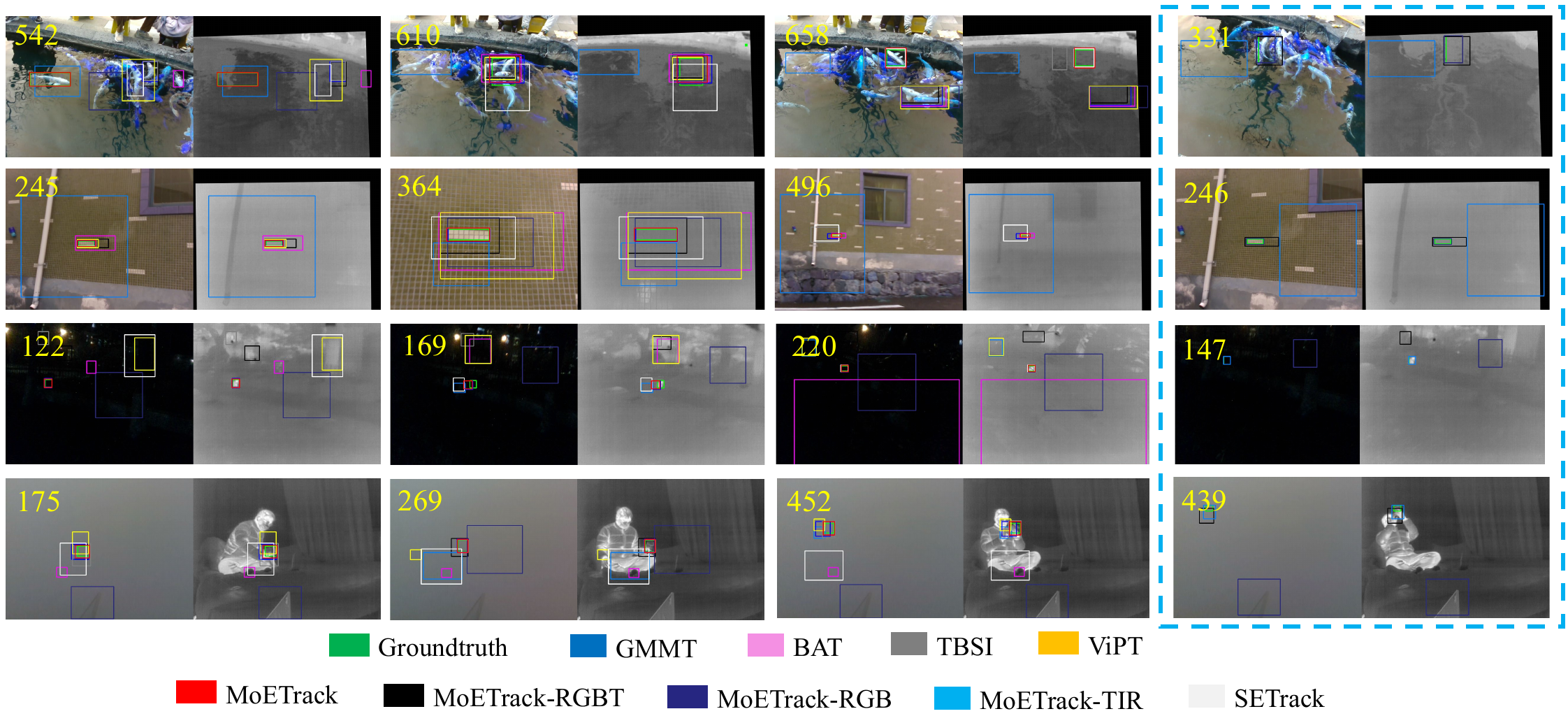}
  \caption{Visualisations on MV-RGBT. From top to bottom, the frames are sampled from \textit{ET\_Fish\_River3}, \textit{ET\_Sign\_Wall1}, \textit{ER\_Cat\_Lawn1}, and \textit{ER\_Bottle\_Bedroom}. Additionally, a straightforward comparison among the RGB, TIR, and RGBT experts is provided on the right side.}
  \label{fig:vis}
\end{figure*}

In addition, visualisations of tracking results are provided in Figure \ref{fig:vis}, which intuitively explain the superiority of our approach.

\subsection{Ablation Study}
\begin{figure}[t]
  \centering
  \includegraphics[width=0.5\textwidth]{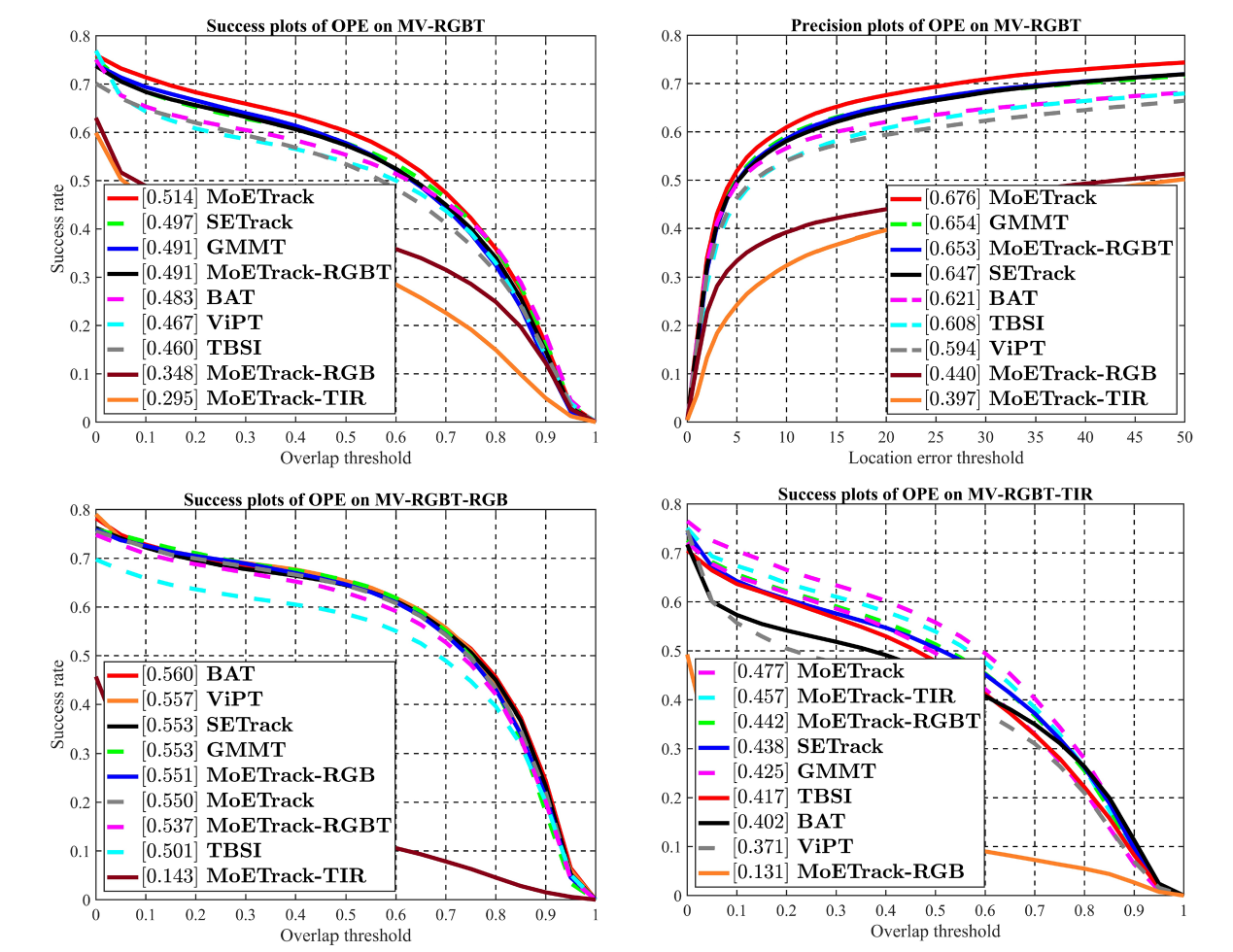}
  \caption{Qualitative analysis on MV-RGBT. }
  \label{fig:results}
\end{figure}

Table \ref{tab:results} and Figure \ref{fig:results} report the ablation study of combining multiple experts.
Basically, consistent with other methods, the variant with only the fused branch is involved, dubbed as SETrack.
Furthermore, in our method, the performance of each expert is also provided.
The experts for RGB, TIR, RGBT branches are referred to as MoETrack-RGB, MoETrack-TIR, and MoETrack-RGBT, respectively.

Firstly, through the comparison between SETrack and MoETrack, continuous improvements can be found on all benchmarks, which strongly demonstrates the superiority of our method.
In addition, utilising the results from the same branch, MoETrack-RGBT also exceeds SETrack on all benchmarks.
This is attributed to the extra losses for the RGB and TIR branches. 
In this way, enhanced RGB and TIR features can be obtained, which further produces boosted fused features for the RGBT expert, leading to improved performance.
However, on MV-RGBT, MoETrack-RGBT performs slightly worse than SETrack on SR but better on PR.
This is mainly because MV-RGBT focuses more on the timing of fusion, which makes the enhanced RGB and TIR features less significant.
It also confirms that MV-RGBT concentrates more on modality validity, which is consistent with our motivation. 

Besides, compared to other variants, MoETrack-RGB and MoETrack-TIR always show significantly worse performance than SETrack and MoETrack-RGBT, which verifies the superiority of integrating multi-modality data for RGBT tracking.

\subsection{Self Analysis}
Since the superiority of the proposed method has been demonstrated, more analysis of the proposed benchmark, MV-RGBT, and the new problem \textit{when to fuse} are provided in this section.

\begin{table}[t]
  \caption{Qualitative analysis of the benchmarks.}
  \centering \label{tab:balanced-pr}
  \resizebox{1\linewidth}{!}{
\begin{tabular}{cccccc}
\toprule
PR/\%           & GTOT & RGBT234 & LasHeR & VTUAV-ST & MV-RGBT \\
\midrule
MoETrack-RGBT   & 92.9 & 87.5    & 71.7   & 82.9     & 65.3    \\
MoETrack-RGB    & 84.9 & 81.6    & 62.4   & 76.1     & 44.0    \\
MoETrack-TIR    & 64.3 & 76.5    & 59.8   & 51.7     & 39.7    \\
\midrule
(1-TIR/RGBT)/\% $\uparrow$ & 30.8 (3) & 12.6 (5)   & 16.6 (4)   & 37.4 (2)     & 39.3 (1)    \\
(1-TIR/RGB)/\% $\downarrow$ & 24.3 (4) & 6.2 (2)    & 4.2 (1)    & 32.1 (5)    & 9.8 (3)     \\    
mRank $\downarrow$        & 3.5  & 3.5     & 2.5    & 3.5      & 2      \\
\bottomrule
\end{tabular}
}
\end{table}

\begin{table}[t]
  \caption{Quantitative results on GTOT, RGBT234, LasHeR, and VTUAV-ST benchmarks.
  }
  \centering \label{tab:balanced-sr}
  \resizebox{1\linewidth}{!}{
\begin{tabular}{cccccc}
\toprule
SR/\%           & GTOT & RGBT234 & LasHeR & VTUAV-ST & MV-RGBT \\
\midrule
MoETrack-RGBT   & 77.7 & 64.8    & 57.5   & 69.1     & 49.1    \\
MoETrack-RGB    & 68.9 & 60.7    & 50.2   & 65.7     & 34.8    \\
MoETrack-TIR    & 56.3 & 54.0    & 47.4   & 41.2     & 29.5    \\
(1-TIR/RGBT)/\% $\uparrow$ & 27.6 & 16.7    & 17.6   & 40.4     & 40.0    \\
(1-TIR/RGB)/\% $\downarrow$ & 18.3 & 11.1     & 5.6    & 37.4     & 15.3     \\
mRank $\downarrow$        & 3.5  & 3.5     & 2.5    & 3.5      & 2.5      \\
\bottomrule
\end{tabular}
}
\end{table}

\textbf{Significance of MV-RGBT:}
This is introduced both quantitatively and qualitatively.

Quantitatively, the statistics displayed in Table \ref{tab:benchmarks} show that MV-RGBT is the most diverse benchmark, encompassing the largest number of object categories and scenes.
Additionally, observations from Table \ref{tab:results} and Figure \ref{fig:results} indicate that the tracking performance on our benchmark is notably lower than that on other benchmarks. 
This suggests that MV-RGBT presents greater challenges than existing benchmarks, thereby fostering the advancement of RGBT tracking.

Qualitatively, Table \ref{tab:balanced-pr} presents the gap between the worst single-modal (TIR) tracker and the multi-modality (RGBT) tracker, as well as the gap between RGB and TIR trackers in terms of PR.
Generally, a larger score for the former indicates that the benchmark can better showcase the significance of aggregating multi-modality information, while a lower score for the latter suggests that different modalities are more balanced. 
Based on these observations, an averaged ranking, mRank, is computed as a comprehensive indicator.
Through the analysis on PR, MV-RGBT ranks first and a similar analysis on SR is provided in the Table \ref{tab:balanced-sr}, where MV-RGBT and LasHeR are equally measured as the best.
Therefore, in terms of the joint evaluation of modality balance and multi-modality significance, MV-RGBT emerges as the most balanced benchmark, showing great potential to accelerate the research in RGBT tracking.

Besides, according to the challenges introduced in Sec. \ref{sec:collection}, MV-RGBT can be further divided into two parts, MV-RGBT-RGB and MV-RGBT-TIR.
Videos suffering extreme illumination and bad weather belong to the second part, as the effectiveness of the RGB modality is critically influenced, while the remaining videos form the first part. 
This implies that each part has different dominating modalities, allowing for a new approach to analysis, which is discussed in Sec. \ref{sec:modalitybias}.

\textbf{When to Fuse:}
As for the discussion on \textit{when to fuse}, it is crucial to first clarify why it should be highlighted.
Accordingly, the results obtained on MV-RGBT-RGB and MV-RGBT-TIR are shown in the lower part of Figure \ref{fig:results}, with a focus on the variants MoETrack-RGBT, MoETrack-RGB, and MoETrack-TIR.
On MV-RGBT-RGB, MoETrack-RGB shows better performance than MoETrack-RGBT, indicating that fusion may not be necessary since MoETrack-TIR performs poorly.
Additionally, on MV-RGBT-TIR, MoETrack-TIR achieves the best result while MoETrack-RGB performs the worst.
These findings reflect that compared to a single modality, the fusion of multi-modality information is not always beneficial, prompting a further discussion on \textit{when to fuse}. 

\begin{figure}[t]
  \centering
  \includegraphics[width=0.45\textwidth]{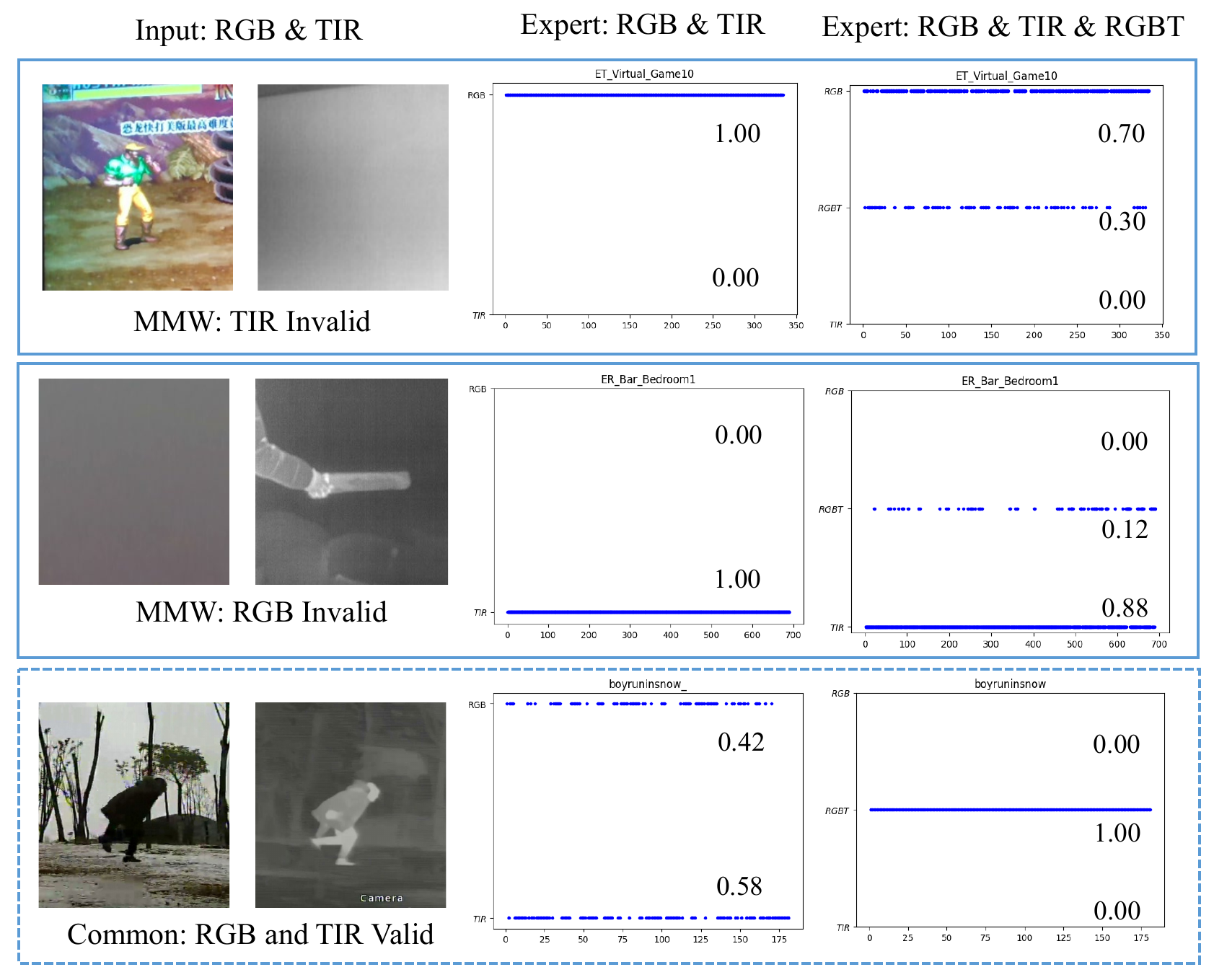}
  \caption{Analysis for proposed new problem  \textit{when to fuse}. }
  \label{fig:experts}
\end{figure}

In this paper, based on the combination of multiple experts, the discussion of \textit{when to fuse} is transferred to the selection among these experts.
Fusion is considered necessary if the results from the RGBT expert are chosen, and vice versa.
Figure \ref{fig:experts} shows the selection results on three videos, visualising the choice in each frame and the ratio of selected frames for each expert.
In the second example from MV-RGBT, where the RGB modality is affected by heavy fog, leading to the object being unseen in the RGB data, the TIR tracking results are consistently more reliable than those obtained by the RGB expert throughout the sequence. 
After involving the RGBT expert,  the selection switches to the RGBT expert for 12\% of all the frames. 
This is attributed to that the offline-trained and parameter-fixed model makes it hard to have the correct choice for every frame.
Nevertheless, results from the TIR expert are predominantly chosen, indicating that multi-modality information fusion may be unnecessary in MMW scenarios.
The same conclusion can be drawn from the first example.
On the contrary, in the third example from common scenarios, the reliabilities of the RGB and TIR experts are nearly the same throughout the video (0.42 \& 0.58), indicating a slight gap between these two experts. 
This further indicates that both modalities are informative for the tracking task. 
After combination, the RGBT branch has further enhanced features, which explains the phenomenon that the results from the RGBT expert dominate the decision in this video.
It means integrating multi-modality information in common scenarios is helpful, as the features from different modalities can mutually reinforce each other for performance boosting.

In conclusion, densely applying the multi-modality fusion proves beneficial in common scenarios, where the data of existing benchmarks is collected.
However, in MMW scenarios, where a single modality alone may not adequately support the tracking task in practical applications, indiscriminate fusion may not only be unhelpful but could also prove detrimental, as indicated by our results.

\subsection{Compositional Analysis for Algorithms}\label{sec:modalitybias}
The proposed benchmark, MV-RGBT, can be stratified into two parts, MV-RGBT-RGB and MV-RGBT-TIR. 
In the former, data predominantly relies on the RGB modality, while the latter exhibits higher-quality data for the TIR modality.
This stratification motivates us to conduct a compositional analysis, evaluating the performance of methods on RGB and TIR subsets separately.
Figure \ref{fig:results} presents the overall (upper figures) and 
partial results (lower figures) at the same time.
On MV-RGBT-RGB, BAT \cite{bat} and ViPT \cite{vipt} outperform  MoETrack, SETrack, and GMMT \cite{gmmt}.
However, their performance drastically deteriorates on MV-RGBT-TIR, only better than MoETrack-RGB.
In contrast, MoETrack, SETrack, and GMMT have a more balanced performance across both MV-RGBT-RGB and MV-RGBT-TIR subsets, thus explaining their overall excellence in evaluation. 
Furthermore, the superior performance of MoETrack, SETrack, and GMMT underscores the importance of a balanced design for each modality, suggesting a promising direction for future studies.

\subsection{Efficiency Analysis}
\label{sec:efficiency}
The efficiency analysis is provided in Table \ref{tab:results}, revealing that an optimal balance between the performance and computational efficiency is exhibited in our method.
Compared to ViPT, our method consumes more time (23 FPS), which is owed to applying the complicated transformer architecture to both RGB and TIR branches.
However, our method consistently outperforms ViPT across all the benchmarks.
Moreover, when compared to other state-of-the-art trackers such as GMMT and BAT, our method demonstrates superior efficiency while maintaining better performance.

Besides, adopting multiple experts brings a slight decrease in efficiency,  albeit with consistent improvements across all benchmarks.
In our design, compared to SETrack, two CNN-based tracking heads and a comparison of the confidence scores are executed in extra.
The former is lightweight and therefore causes a slight reduction in efficiency ($\Delta$=2FPS).
The latter expends neglectable time since it only involves fetching the maximum value among three scalars.
Despite this minor efficiency trade-off, the adoption of multiple experts leads to improved performance across all the benchmarks, with particularly notable enhancements observed on MV-RGBT. 

\subsection{Beyond RGBT Tracking}\label{sec:dicsussion}
Basically, one of the key contributions of this work lies in the demonstration that fusion is not always necessary for multi-modality tasks and a detailed discussion is carried out on RGBT tracking.
However, our insight is not limited to a specific area and has a broader applicability beyond RGBT tracking.
It can be extended to various multi-modality tasks, such as RGBD/RGBE tracking and RGBT detection.
Moreover, by leveraging the benchmark proposed in this work, researchers can directly conduct comprehensive evaluations and analyses to ascertain the efficacy of fusion strategies in RGBT detection, which is supposed to facilitate the development of more robust multi-modality detection systems.
The stability of video analysis under degraded scenarios can be further analysed \cite{wu2023motion, chen2021hybrid}.

\section{Conclusion}
With the awareness of the inconsistency between the existing benchmarks and the multi-modality warranting (MMW) scenarios where the advantages of multi-modality information are most pronounced, we presente a diverse and challenging benchmark, namely MV-RGBT, by ensuring all the data in MMW scenarios.
In this way, the inconsistency is removed and the evaluation in MMW scenarios can be executed, thereby providing more reliable suggestions for the deployment of RGBT trackers in practical applications.
Besides, a further division of MV-RGBT enables a compositional analysis of existing methods, revealing the advantages of multi-modality balanced designs for achieving higher overall performance in RGBT tracking.

Additionally, inspired by widely-appeared invalid data in MMW scenarios, a new problem \textit{when to fuse} is posed and discussed by devising a new MoETrack method with multiple experts.
Through building new state-of-the-art records on RGBT234, LasHeR, VTUAV-ST, and MV-RGBT, the superiority of MoETrack is demonstrated and the extensive experiments also indicates that when the information in both modalities are of good quality, the fused results are always more reliable.
On the contrary,  when one modality contains non-informative data, fusion not only has a large potential to be unnecessary but could also degrade performance.

\section{Acknowledgements}
This work was supported by the National Key Research and Development Program of China (Grant No. 2023YFF1105102, 2023YFF1105105), the National Natural Science Foundation of China (Grant NO. 62020106012, 62332008, 62106089, U1836218, 62336004), the 111 Project of Ministry of Education of China (Grant No.B12018), and the UK EPSRC (EP/N007743/1, MURI/EPSRC/DSTL, EP/R018456/1).
\bibliographystyle{IEEEtran}
\bibliography{ref}

\clearpage
\newpage

\end{document}